\documentclass[journal]{IEEEtran}

\ifCLASSINFOpdf
\else
   \usepackage[dvips]{graphicx}
\fi
\usepackage{url}

\usepackage{graphicx}
\usepackage{subfigure}
\usepackage{gensymb}     
\usepackage{array}
\usepackage{amsmath,amssymb}    

\newcommand{\RNum}[1]{\uppercase\expandafter{\romannumeral #1\relax}}   

\begin{document}
\title{Stereo-based Multi-motion Visual Odometry for Mobile Robots}

\author{Qing Zhao, Bin Luo, and Yun Zhang
\thanks{The authors are with the State Key Laboratory of Information Engineering in Surveying, Mapping and Remote Sensing, Wuhan University, Wuhan 430072, China (e-mail: zhaoqing@whu.edu.cn; luob@whu.edu.cn; zhangyunmail@whu.edu.cn).}}
\maketitle

\begin{abstract}
With the development of computer vision, visual odometry is adopted by more and more mobile robots. However, we found that not only its own pose, but the poses of other moving objects are also crucial for the decision of the robot. In addition, the visual odometry will be greatly disturbed when a significant moving object appears. In this letter, a stereo-based multi-motion visual odometry method is proposed to acquire the poses of the robot and other moving objects.  In order to obtain the poses simultaneously, a continuous motion segmentation module and a coordinate conversion module are applied to the traditional visual odometry pipeline.  As a result,  poses of all moving objects can be acquired and transformed into the ground coordinate system.  The experimental results show that the proposed multi-motion visual odometry can effectively eliminate the influence of moving objects on the visual odometry,  as well as achieve 10 cm in position and 3\degree \ in orientation RMSE (Root Mean Square Error) of each moving object.
\end{abstract}


\IEEEpeerreviewmaketitle

\section{Introduction}
\IEEEPARstart{W}{ith} the development of computer vision and the emergence of various high-performance vision sensors, visual odometry has been widely used in unmanned driving, intelligent robots, unmanned aerial vehicles and other fields. Visual odometry is the process of estimating the position and orientation of a robot by analyzing the associated camera images \cite{Nister2004Visual}, and advantages have been shown in the scene where the motion information from other sources is difficult to obtain, especially in the outer space \cite{Aqel2016Review}. At present, visual odometry is mainly divided into direct method and feature based method. Pixel intensity in the image is directly used as the input by the direct method, based on the assumption that the light intensity is unchanged in the image sequence. LSD-SLAM \cite{Engel2014LSD} and DSO \cite{Engel2016Direct} are two classical algorithms of the direct method. PTAM \cite{Klein2007Parallel} and ORB-SLAM \cite{Mur2017ORB} are two excellent algorithms of the feature based method, which extract the features from the image firstly and then estimate the pose of the robot by the features. Both two methods can obtain accurate pose estimation in a relatively static environment, however, there are often some significantly moving objects in the real scene, such as other vehicles in the automatic driving scene and pedestrians in the mobile robot scene, both methods will be seriously affected in such dynamic cases.

In order to apply visual odometry to real applications, many methods have been propose. For example, \cite{Sahdev2018Indoor} uses wheeled odometry combined with visual odometry to eliminate the influence of dynamic scene, \cite{Kim2017Effective} determines the feature points belonging to the static background by estimating the background model. By eliminating the conflict points, lots of methods e.g., \cite{Zhi2008Stereo} and \cite{Azartash2014Fast} can often obtain relatively robust pose estimation in the dynamic scene, but also lose a lot of important information, such as the trajectory and motion trend of the moving objects, which is also crucial for the decision-making of the robot.

Recently, many researches have been carried out to obtain the visual odometry of the robot and other moving objects based on multi-motion segmentation. Optical flow \cite{Horn1980Determining} is used to find the velocities of the points in the scene and points with similar velocities are considered to be the same moving object. Appearance-based tracking techniques \cite{Milan2016MOT16} detect objects in images and then solve the motion estimation, but hardly to handle detection errors or unknown types. Subspace techniques cluster sparse feature points and their motions into lower-dimensional subspaces using the specified camera models, this technique has been applied in \cite{Costeira1998A} to classify mutiple bodies where points may belong to different moving objects while still remain very sensitive to noise. In order to obtain the $SE(3)$ pose of each moving object, the energy minimization method is applied in \cite{Judd2018Multimotion} to segment the stereo camera observations. The energy minimization method in multi-motion estimation can obtain the labels of objects and their corresponding motion models simultaneously by setting the cost functions and continuously optimizing the error. The cost functions describe how well the estimated trajectories conform to the labeled objects, e.g., reprojection error and piecewise smoothness. It should be noted that this approach requires a lot of frames for a complex initialization.

In the last few years, a great many sampling based multi-model fitting methods, e.g., \cite{Tennakoon2018Effective}, \cite{Barath2019Progressive} and \cite{Xun2018Motion} have been proposed for multi-motion segmentation. It is efficent to generate a large number of initial model hypotheses and models with largely overlapping inlier sets are merged. Models with the largest nonoverlapping inlier sets after merging are taken as the different objects. In this letter, a multi-motion visual odometry pipeline is designed to acquire the visual odometry of multiple moving objects. Compared with the traditional visual odometry, a multi-model fitting based continuous motion segmentation module and a coordinate transformation module are added in the multi-motion visual odometry pipeline.

\begin{figure*}[htbp]
\centering
\includegraphics[width=\linewidth]{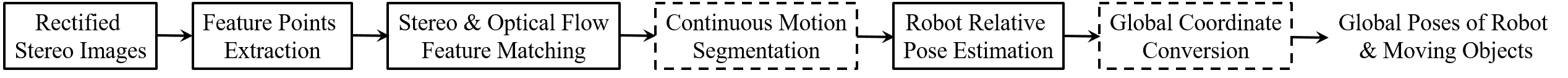}
\caption{An illustration of the proposed multi-motion visual odometry pipeline, two new modules are included in the dotted boxes.}
\end{figure*}
\section{Methodology}
Since multi-model fitting algorithm has been proved to be able to cluster feature points belonging to various models and estimate parameters of each model, it is widely used in motion segmentation with multiple moving objects. Methods such as J-linkage \cite{Toldo2008Robust},\cite{Magri2014T}, AKWSH \cite{Magri2016Multiple}, and T-Linkage \cite{Hanzi2012Simultaneously} are considered to be classic multi-model fitting algorithms. In this letter, a multi-model fitting method based on permutation residuals \cite{Yun2017Permutation} is adopted, which can effectively increase the variance between classes and decrease the variance within classes. It can also realize the adaptive adjustment of various threshold values, making the moving objects with different motion models more separable.

Although the multi-model fitting algorithm can classify the inlier points of models belonging to different moving objects, the given classification labels are always random and discontinuous, so it cannot be directly used in continuous image sequences. Therefore, a continuous motion segmentation module and a coordinate conversion module are designed and applied into the traditional pipeline of visual odometry (Fig. 1).

\subsection{Continuous Motion Segmentation Module}
The motion segmentation module is mainly consisted of two parts:
\subsubsection{Combination of Key Frames and Normal Frames}
Since the low speed and high frame rate characteristics of mobile robots, the motion segmentation in frame-by-frame mode is always unsatisfactory. Therefore, the concept of key frame in visual slam is used for reference. The criteria for selecting key frames are as follows: a) certain time interval; b) the number of matching points with the previous key frame is less than a certain threshold. The motion in adjacent key frames is relatively obvious, which is more suitable for motion segmentation, and at the same time, it can reduce a lot of computational work and improve the segmentation efficiency. For normal frames, they are matched with the nearest previous key frame, the matched feature points are assigned with the corresponding labels and the rest feature points are considered as new points and the k-nearest neighbor method is used in the assignment of the label after triangulation.
\subsubsection{Tracking the Labels of Motion Models}
In order to get the poses of the robot and each moving object, it is essential to assign each moving object with an accurate and stable label throughout the whole moving process. However, due to the inescapable mismatching of feature points, especially when the robot moves too violently, only associating labels between two adjacent key frames may lead to degradation. Therefore, we propose a joint label association method based on sliding window. Firstly, the current frame t$_k$ and the previous 4 adjacent key frames are specified as a sliding window. Then, t$_k$ is matched with t$_{k-1}$ to t$_{k-4}$ respectively and the labels are passed from the feature points to the matched points. After this, for each matching point, the labels passed from each key frame in the sliding window are weighted and counted, and the highest proportion label is assigned to obtain a relatively stable motion segmentation label,
\begin{equation}
{S(t_k) = \sum_{i=1}^{n}l(i) \times w(i)},
\end{equation}
where $n$ is one less than the length of the sliding window, $l(i)$ represents the different labels obtained by feature matching and $w(i)$ represents the weights corresponding to different key frames, which  decrease as the time interval increases, $S(t_k)$ is a polynomial made up of labels and corresponding coefficients at $t_k$, e.g., $S(t_k) = 0.2Label\_a + 0.4Label\_b + 0.6Label\_c + 0.8Label\_d$. The label with the maximum coefficient will be selected, however, if the coefficient is less than a certain threshold, it is considered to be a new moving object.
\subsection{Coordinate Conversion Module}
Through the motion segmentation module, the feature points belonging to each moving object are respectively brought into the traditional visual odometry pipeline, and the poses of robot are obtained relative to each moving object. However, the pose of each moving object in the ground coordinate system is considered to be the final output, so the coordinate conversion module is crucial. In this module, obtaining the pose of each moving object relative to the robot at the initial moment is considered to be most difficult. Firstly, the traditional visual odometry method is used to perform pose estimation on points belonging to each motion model obtained after motion segmentation. Then, for each moving object, the triangulated feature points are projected into the first frame according to the estimated pose, and the surface point set of the moving object is obtained,
\begin{equation}
{\textbf{P}_{obj} = \textbf{P}_{cam}\textbf{T}^{-1}},
\end{equation}
where $\textbf{P}_{cam}$ and $\textbf{P}_{obj}$ are homogeneous coordinates in the form of row vectors, $\textbf{P}_{cam}$ represents the triangulated 3d coordinates of a feature point in the current frame and $\textbf{P}_{obj}$ represents the 3d coordinates of the point after being projected to the first frame, $T$ equals to $\bigl( \begin{smallmatrix} \textbf{R} & \textbf{t} \\ \textbf{0} & \textbf{1} \end{smallmatrix} \bigr)$, which is the homogeneous pose of the robot relative to the corresponding moving object in the current frame. As the number of observation frames increases, the number of points on the surface of each moving object increases. Next, the space of the surface point set is evenly divided, and the grid holding at least one surface point is considered to be occupied. In this way, the weight of the newcome edge point is increased to be the same as the other points. Finally, the gravity center of the occupied grids is obtained and the result is considered to be the initial position of each moving object relative to the robot. The pose of each moving object in the ground coordinate system can be obtained by
\begin{equation}
{\textbf{T}_{move\rightarrow ground} = \textbf{T}_{robot}\textbf{T}_{robot\rightarrow move}^{-1}\textbf{T}_{init}^{-1}},
\end{equation}
where $\textbf{T}_{move\rightarrow ground}$ is the pose of the moving object in the ground coordinate system, $\textbf{T}_{robot}$ is the  pose of the robot, $\textbf{T}_{robot\rightarrow move}$ represents the pose of the robot relative to the moving object and $\textbf{T}_{init}$ represents the initial pose of the moving object relative to the robot, that is, the gravity center of the surface point set of the moving object, obtained by formula (2).
\section{Experiment}
The performance of the stereo-based multi-motion visual odometry algorithm is tested on a real-world dataset collected by our differential robot platform, which is equipped with a Zed stereo camera and a Hokuyo 2D lidar with a range of 30 meters(Fig. 2). Unlike other multi-motion segmentation and tracking datasets, our data is mainly collected in a public area of an office building with abort 50 meters long and 10 meters wide and the trajectories of the robot and other moving objects are consisted of L-shaped and S-shaped, with the longest reaching about 35 meters, which can better reflect the working state of the robot in the real environment(Fig. 3). The ground truth poses of the robot and other moving objects are obtained by particle filter \cite{Lei2012Self}, which matches the two-dimensional laser data with the prior laser mapping result \cite{Hess2016Real} of the public area. It is proved that particle filter has the ability to maintain the accuracy of the pose even when the environment changes a lot. After being compared with the actual manual measurements, the accuracy of the map is considered to be within 5cm and the precision of the poses reaches 5cm in position and 1\degree \ in orientation.

\begin{figure}[htbp]
\centering
\subfigure[Zed Camera]{
\begin{minipage}[t]{0.5\linewidth}
\centering
\includegraphics[height=1in]{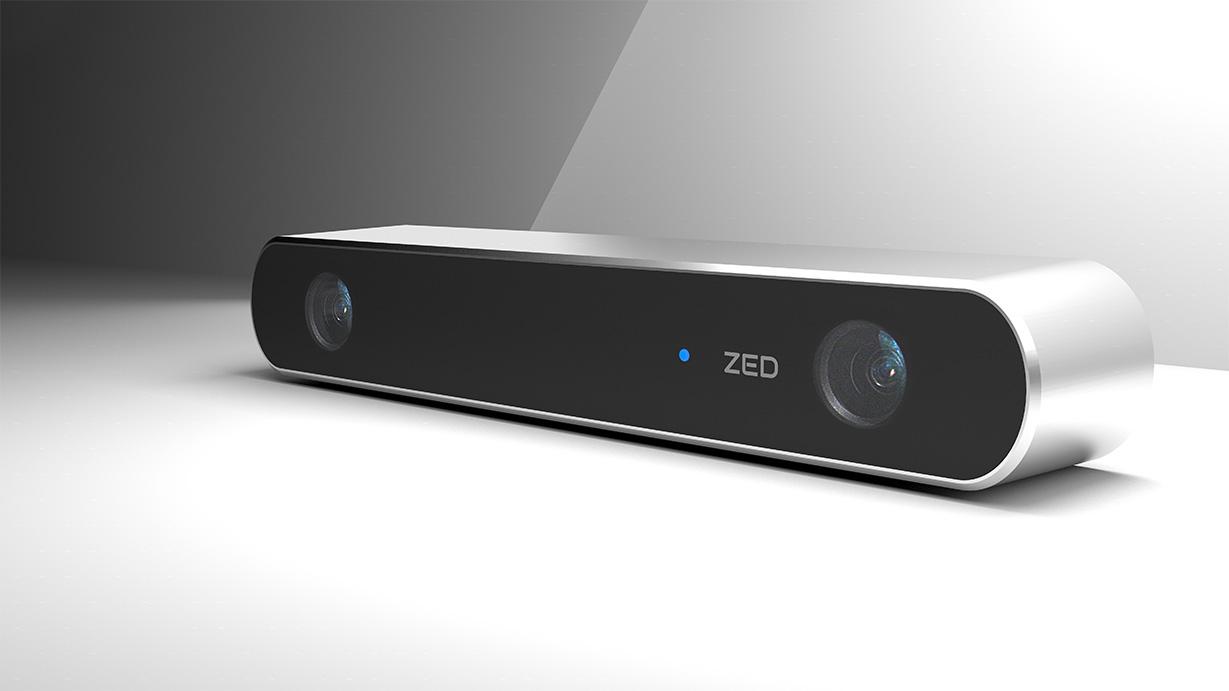}
\end{minipage}
}%
\subfigure[Hokuyo lidar]{
\begin{minipage}[t]{0.5\linewidth}
\centering
\includegraphics[height=1in]{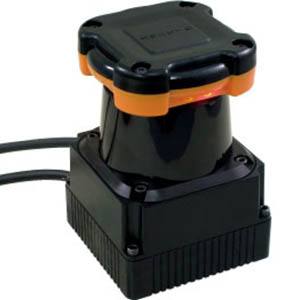}
\end{minipage}
}
\caption{(a) has the ability to record 1080p stereo video at 30 fps and its baseline is 12cm, (b) is a two-dimensional lidar with a range of 30 meters, a scanning angle range of 270\degree \ and a rotation rate of 40 times per second}
\end{figure}

\begin{figure}[htbp]
\centering
\subfigure[map]{
\begin{minipage}[t]{0.5\linewidth}
\centering
\includegraphics[height=1in]{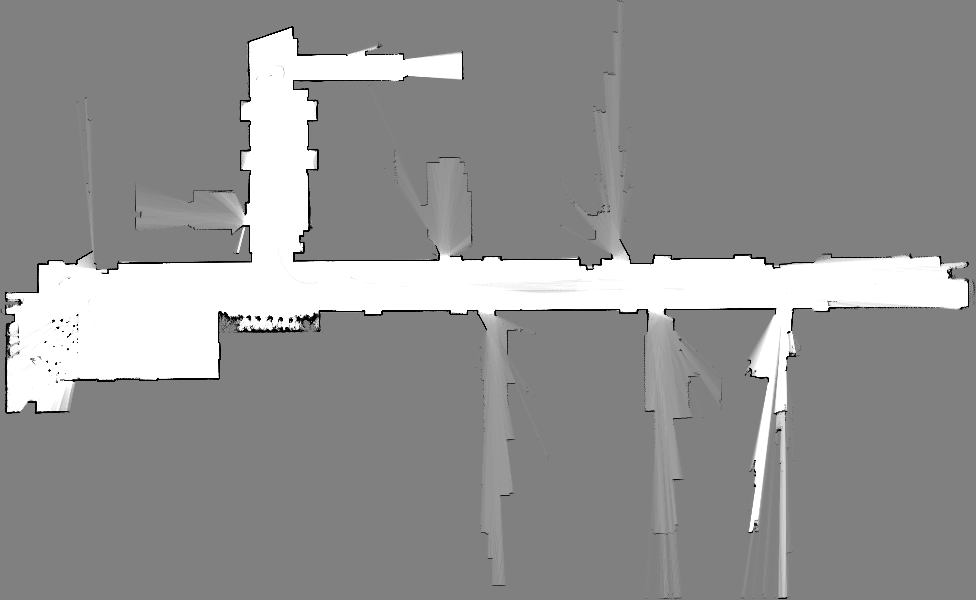}
\end{minipage}
}%
\subfigure[trajectory]{
\begin{minipage}[t]{0.5\linewidth}
\centering
\includegraphics[height=1in]{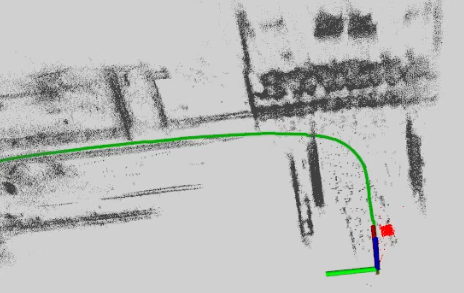}
\end{minipage}
}
\caption{(a) is a grid map constructed from two-dimensional laser data using the cartographer algorithm , (b) shows us a L-shaped trajectory}
\end{figure}

The experimental results were produced by 5 image sequences in total, including L-shaped and S-shaped trajectories with static background and a significant moving object respectively, and a forward and backward movement with two significant moving objects. Estimation was performed by a certain group of parameters, with 5 continuous frames for the time interval of key frame, 500 matched points with the previous key frame as the threshold to select a key frame, 300 matched points to adjust the extraction scale of feature points. Feature detection and matching were performed using LIBVISO2 \cite{Geiger2012StereoScan} and the pose estimation was performed with ORB-SLAM2 pipeline(without loop closure) in stereo mode. In Fig. 4, there is one robot and two moving objects in the scene and the effect of continuous frames motion segmentation according to the method in the previous section is shown. It is reflected that the label tracking is stable, the number of the motion models and the segmentation of objects with different motion models are accurate.

\begin{figure}[htbp]
\centering
\subfigure[]{
\begin{minipage}[t]{0.5\linewidth}

\includegraphics[width=1.6in]{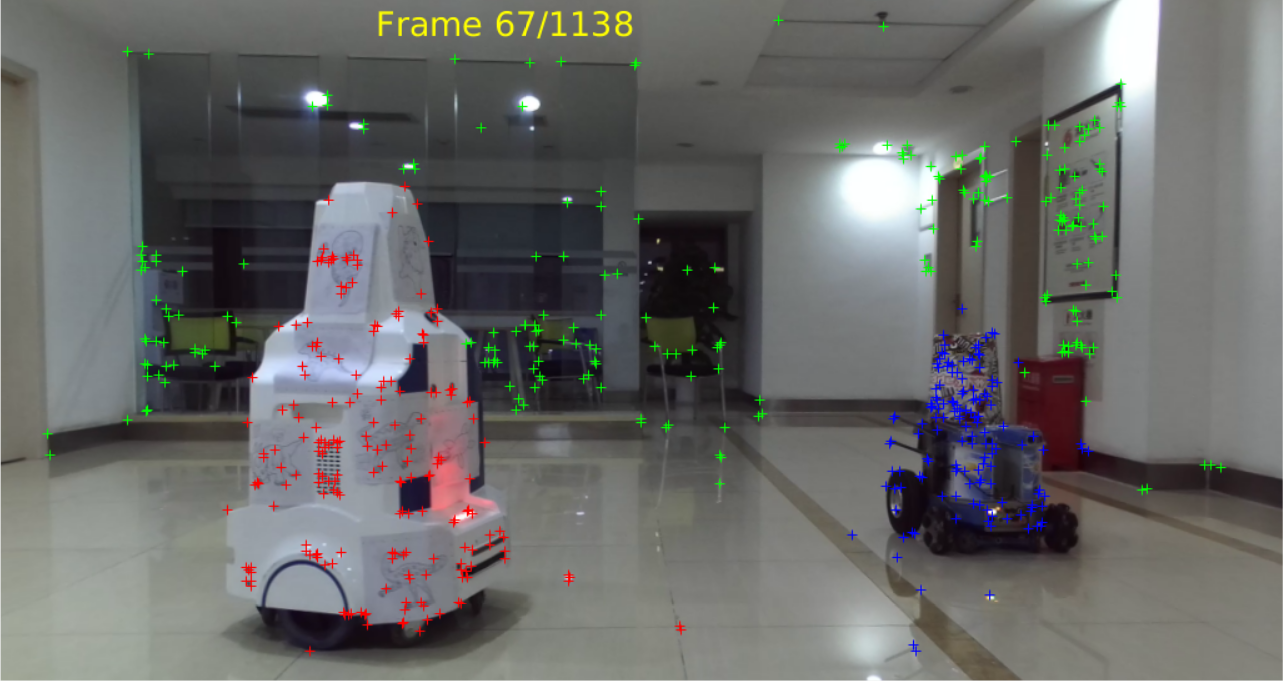}
\end{minipage}
}%
\subfigure[]{
\begin{minipage}[t]{0.5\linewidth}

\includegraphics[width=1.6in]{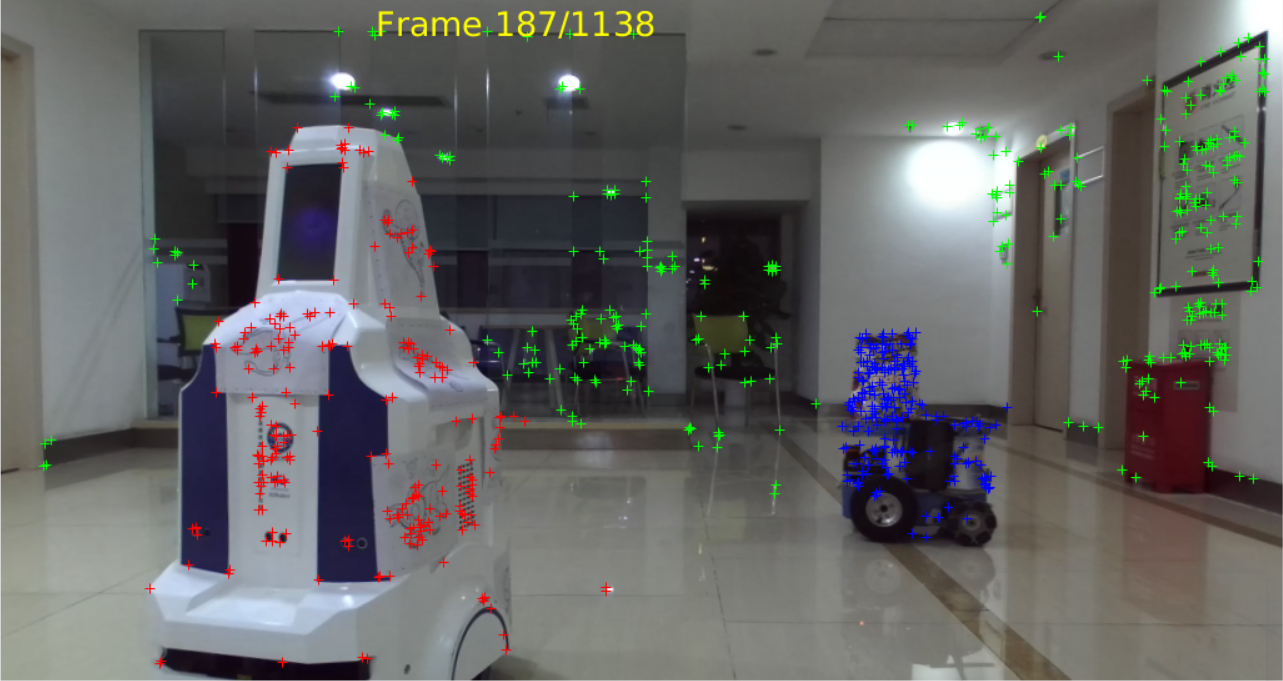}
\end{minipage}
}
\subfigure[]{
\begin{minipage}[t]{0.5\linewidth}

\includegraphics[width=1.6in]{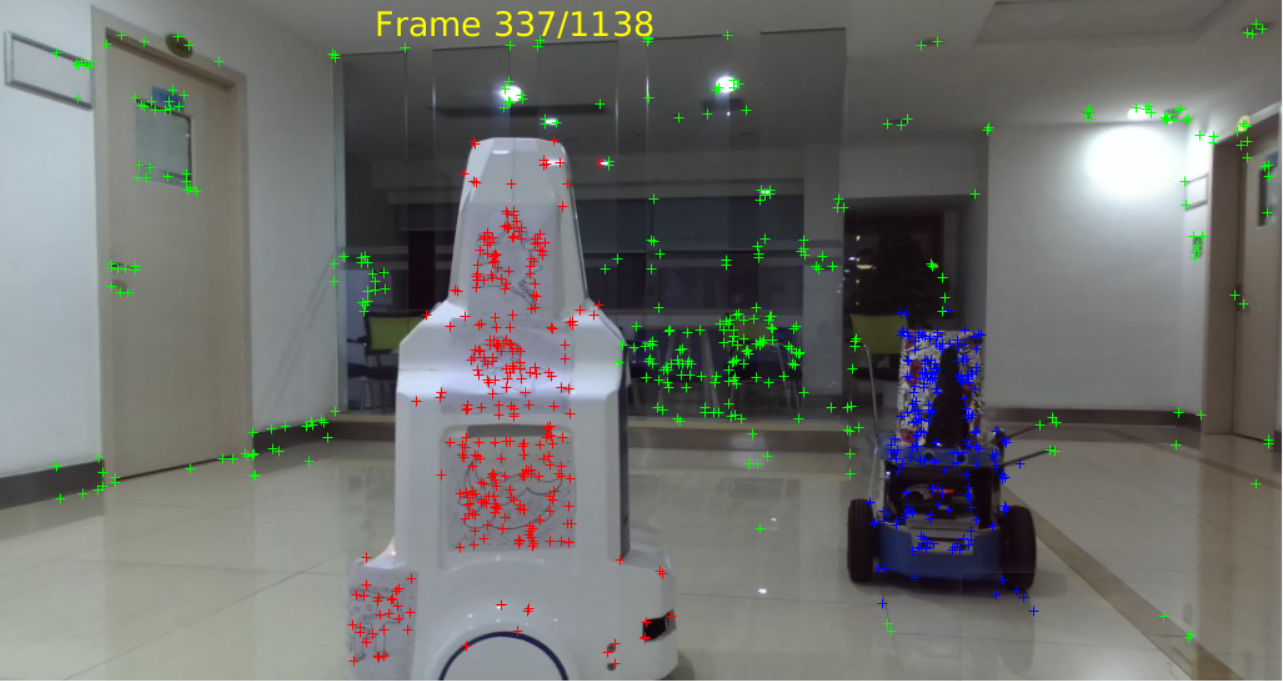}
\end{minipage}
}%
\subfigure[]{
\begin{minipage}[t]{0.5\linewidth}

\includegraphics[width=1.6in]{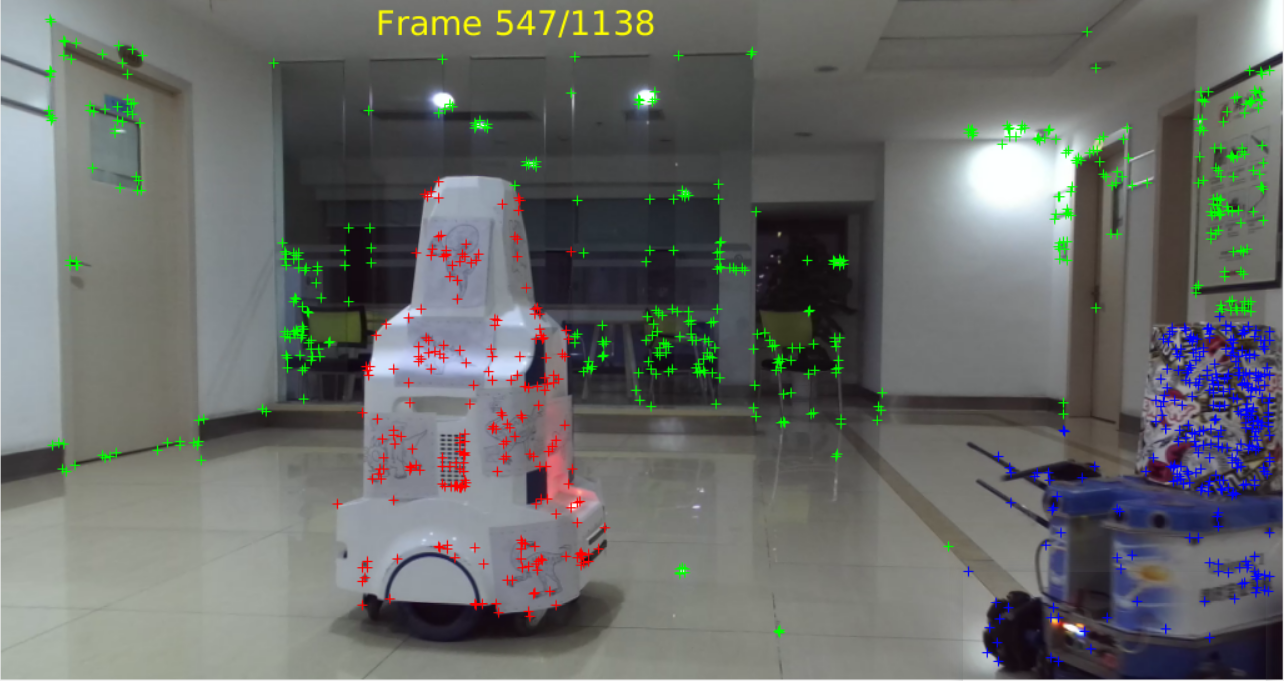}
\end{minipage}
}
\caption{In this scene, there is a robot moving forward and backward, a moving object moving along a square track and another moving object rotating in place. The 4 subgraphs show the results of motion segmentation at 4 moments. The points in different colors represent objects belonging to different motion models(the green points represent the background)}
\end{figure}

Fig. 5 shows the result of pose estimation for 5 image sequences respectively. In order to make the presentation more clear, a 1:30 extraction of the estimated pose is conducted  for visualization. In (a) and (b), the robot moves on L-shaped and S-shaped trajectory respectively in the static background, without other significant moving objects and the traditional visual odometry method is used to estimate the pose of the robot. (c) and (d) belong to the same image sequence, the two sets of trajectories are presented separately due to their high overlap. The former represents the trajectory of the robot, while the latter represents the trajectory of the moving object. The poses of the robot and the moving object are estimated by the proposed multi-motion visual odometry method in (c)-(f). There are two other moving objects in (f), one is rotating in situ and not shown in the figure.
\begin{figure*}[htbp]
\centering
\subfigure[]{
\begin{minipage}[t]{0.33\linewidth}
\centering
\includegraphics[width=2.3in]{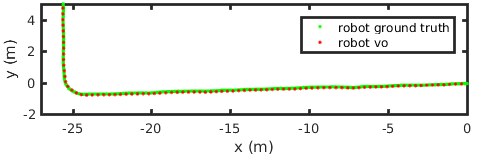}
\end{minipage}
}%
\subfigure[]{
\begin{minipage}[t]{0.33\linewidth}
\centering
\includegraphics[width=2.3in]{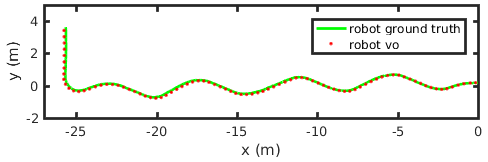}
\end{minipage}
}%
\subfigure[]{
\begin{minipage}[t]{0.33\linewidth}
\centering
\includegraphics[width=2.3in]{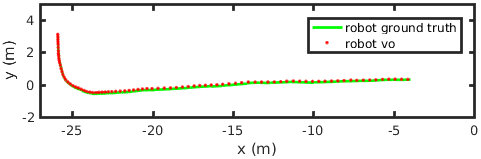}
\end{minipage}
}
\subfigure[]{
\begin{minipage}[t]{0.33\linewidth}
\centering
\includegraphics[width=2.3in]{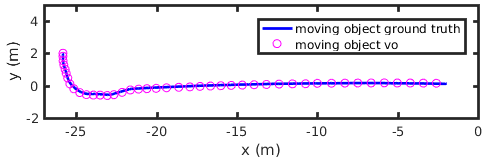}
\end{minipage}
}%
\subfigure[]{
\begin{minipage}[t]{0.33\linewidth}
\centering
\includegraphics[width=2.3in]{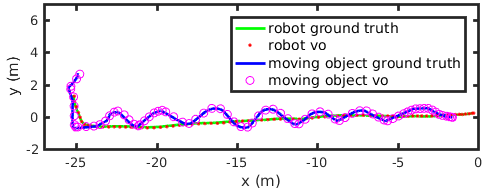}
\end{minipage}
}%
\subfigure[]{
\begin{minipage}[t]{0.33\linewidth}
\centering
\includegraphics[width=2.3in]{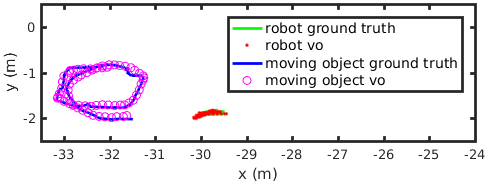}
\end{minipage}
}
\caption{ In (a) and (b), the robot moves on L-shaped and S-shaped trajectory respectively in the static background; (c) represents the trajectory of the robot and (d) represents the trajectory of the moving object in the same sequence; There are two other moving objects in (f), one is rotating in situ and not shown in the figure. In (a) and (b), poses are estimated by the traditional visual odometry method , the multi-motion visual odometry method is used in the other figures.}
\end{figure*}
\begin{table}
\caption{RMSE in position and roatation for 5 image sequences}
\label{table}
\small
\vspace{-1.0em}
\setlength{\tabcolsep}{3pt}
\renewcommand{\arraystretch}{1.4}
\begin{tabular}{p{55pt} p{55pt} p{55pt} p{55pt}}
\hline
\hline
Objects&
Robot&
Object$_1$ &
Object$_2$ \\[1pt]
\hline
\multicolumn{4}{l}{Sequence \RNum{1} (traditional): L-shaped, static background (31m)}\\[1pt]
\hline
Position [m]&0.1084&\quad---&\quad---\\[1pt]
Rotation [\degree]&1.1534&\quad---&\quad---\\[1pt]
\hline
\multicolumn{4}{l}{Sequence \RNum{2} (traditional): S-shaped, static background (35m)}\\[1pt]
\hline
Position [m]&0.1021&\quad---&\quad---\\[1pt]
Rotation [\degree]&2.1034&\quad---&\quad---\\[1pt]
\hline
\multicolumn{4}{l}{Sequence \RNum{3} (traditional): L-shaped, 1 moving object (25m)}\\[1pt]
\hline
Position [m]&FAILED&FAILED&\quad---\\[1pt]
Rotation [\degree]&FAILED&FAILED&\quad---\\[1pt]
\hline
\multicolumn{4}{l}{Sequence \RNum{3} (\textbf{multi-motion}): L-shaped, 1 moving object (25m)}\\[1pt]
\hline
Position [m]&0.1005&0.0572&\quad---\\[1pt]
Rotation [\degree]&1.5233&2.8211&\quad---\\[1pt]
\hline
\multicolumn{4}{l}{Sequence \RNum{4} (traditional): S-shaped, 1 moving object (26m)}\\[1pt]
\hline
Position [m]&FAILED&FAILED&\quad---\\[1pt]
Rotation [\degree]&FAILED&FAILED&\quad---\\[1pt]
\hline
\multicolumn{4}{l}{Sequence \RNum{4} (\textbf{multi-motion}): S-shaped, 1 moving object (26m)}\\[1pt]
\hline
Position [m]&0.0585&0.1123&\quad---\\[1pt]
Rotation [\degree]&1.3554&2.1547&\quad---\\[1pt]
\hline
\multicolumn{4}{l}{Sequence \RNum{5} (\textbf{multi-motion}): 2 moving objects (9m)}\\[1pt]
\hline
Position [m]&0.0661&0.0985&0.0512\\[1pt]
Rotation [\degree]&1.1288&2.3575&1.7623\\[1pt]
\hline
\hline
\end{tabular}
\label{tab1}
\end{table}

The RMSE corresponding to each image sequence in the dataset we collect is presented in Table \RNum{1}, where different motion trajectories, different numbers of moving objects and different kinds of visual odometry methods can be discussed and compared. It can be easily found that when the robot moves in a static environment, the traditional visual odometry method has an excellent performance, however, when significant moving objects appear, the traditional method failed to obtain the pose of the robot. In such scene, the proposed multi-motion visual odometry method is able to be used to obtain accurate pose estimations for both the robot and the moving objects.

Our dataset, in which the track length of four image sequences is more than 25 meters and the number of frames is more than 1,500, is quite long in the related multi-motion datasets. The long moving distance can reflect the real working state of the robot better, and it also brings difficulties in multi-motion segmentation and pose estimation. The position error of the robot and the moving object can be controlled within 0.1 meter, and the orientation error of the moving object is within 3\degree \, while the robot's is better. The proposed multi-motion visual odometry method can simultaneously estimate the pose of robot and multiple moving objects in an advanced level. In the algorithm, the continuous frame motion segmentation module is stable and can maintain stable segmentation label output in the whole process of experiment. Due to the missing of the loop-closure detection module, it will accumulate errors along with the motion. In addition, the blur of image and the inaccurate estimation of the initial relative pose between robot and the moving objects can enlarge the errors to a certain extent.

\section{Conclusion}

In this letter, we have proposed a stereo-based multi-motion visual odometry to obtain the accurate poses of robot and other moving objects at the same time. A continuous motion segmentation module and a coordinate conversion module are applied to the traditional visual odometry pipeline. After being verified on the long sequence dataset we collected, the proposed method is considered to be able to simultaneously estimate the pose of robot and multiple moving objects in an advanced level. The algorithm is a little time consuming and we are trying to optimize the algorithm to achieve a real-time level. When the real-time level is reached, a lot of smart applications can be implemented, such as intelligent obstacle avoidance, intelligent patrol, moving object behavior analysis and so on.

\nocite{*}
\bibliographystyle{unsrt}
\bibliography{ref}



\end{document}